\documentclass[12pt,onecolumn]{IEEEtran}
\usepackage{amsmath,graphicx}
\usepackage{amsfonts}
\usepackage[numbers,sort&compress]{natbib}
\usepackage{algorithmicx}
\usepackage[ruled]{algorithm}
\usepackage{algpseudocode}
\usepackage{caption2}
\usepackage{multirow}
\usepackage{color}
\usepackage{url}

\begin{document}

\def\bsa{{\boldsymbol{a}}}
\def\bsb{{\boldsymbol{b}}}
\def\bsc{{\boldsymbol{c}}}
\def\bsd{{\boldsymbol{d}}}
\def\bse{{\boldsymbol{e}}}
\def\bsf{{\boldsymbol{f}}}
\def\bsg{{\boldsymbol{g}}}
\def\bsh{{\boldsymbol{h}}}
\def\bsi{{\boldsymbol{i}}}
\def\bsj{{\boldsymbol{j}}}
\def\bsk{{\boldsymbol{k}}}
\def\bsl{{\boldsymbol{l}}}
\def\bsm{{\boldsymbol{m}}}
\def\bsn{{\boldsymbol{n}}}
\def\bso{{\boldsymbol{o}}}
\def\bsp{{\boldsymbol{p}}}
\def\bsq{{\boldsymbol{q}}}
\def\bsr{{\boldsymbol{r}}}
\def\bss{{\boldsymbol{s}}}
\def\bst{{\boldsymbol{t}}}
\def\bsu{{\boldsymbol{u}}}
\def\bsv{{\boldsymbol{v}}}
\def\bsw{{\boldsymbol{w}}}
\def\bsx{{\boldsymbol{x}}}
\def\bsy{{\boldsymbol{y}}}
\def\bsz{{\boldsymbol{z}}}

\def\hsx{{\hat{\boldsymbol{x}}}}
\def\hsw{{\hat{\boldsymbol{w}}}}

\def\bsla{{\boldsymbol{\lambda}}}

\def\bszero{{\boldsymbol{0}}}

\title{Reconstruction of Enhanced Ultrasound Images From Compressed Measurements Using Simultaneous Direction Method of Multipliers}

\author{Zhouye Chen, Adrian Basarab, Denis Kouam\'{e}
\thanks{The authors are with University of Toulouse, IRIT CNRS UMR 5505, e-mail: chen,basarab,kouame@irit.fr. 
This journal article is an extended version of the proceedings paper of the
IEEE IUS 2015 conference \cite{Chen2015IUS}.}
}


\maketitle

\begin{abstract}
High resolution ultrasound image reconstruction from a reduced number of measurements is of great interest in ultrasound imaging, since it could enhance both the frame rate and image resolution. Compressive deconvolution, combining compressed sensing and image deconvolution, represents an interesting possibility to consider this challenging task. The model of compressive deconvolution includes, in addition to the compressive sampling matrix, a 2D convolution operator carrying the information on the system point spread function. 
Through this model, the resolution of reconstructed ultrasound images from compressed measurements mainly depends on three aspects: the acquisition setup,  \textit{i.e.} the incoherence of the sampling matrix, the image regularization,  \textit{i.e.} the sparsity prior, and the optimization technique. In this paper, we mainly focused on the last two aspects. We proposed a novel simultaneous direction method of multipliers-based optimization scheme to invert the linear model, including two regularization terms expressing the sparsity of the RF images in a given basis and the generalized Gaussian statistical assumption on tissue reflectivity functions. The performance of the method is evaluated on both simulated and \textit{in vivo} data.
\end{abstract}


\IEEEpeerreviewmaketitle

\section{Introduction}

\IEEEPARstart{S}{ince} the applicability of compressive sampling (CS) to 2D and 3D Ultrasound (US) imaging (e.g. \cite{achim2010compressive,quinsac2012frequency,chernyakova2014fourier, liebgott2012compressive,liebgott2013pre,schiffner2014pulse,david2015time}) or to duplex Doppler \cite{RICH-13} has been proven, the topic of CS in the field of US imaging attracted a growing interest from several research groups. CS is a mathematical framework allowing to recover a compressible image, via non linear optimization routines, from few linear measurements (below the limit standardly imposed by the Shannon-Nyquist theorem) \cite{donoho2006compressed,candes2006robust}. According to the CS theory, this reconstruction is possible provided that the restricted isometry property (RIP), characterizing the measurement matrix, holds \cite{donoho2006compressed,candes2006robust}. RIP has been extensively explored in the literature for several classes of matrices. The most common examples that guarantee the respect of RIP for a number of measurements linearly depending on the sparsity level of the image to recover include random Gaussian or Bernoulli matrices or partial Fourier matrix.     

The interest of CS application in US imaging systems is to increase the frame rate and/or to decrease the amount of acquired data and/or to decrease the computational complexity of beamforming \cite{quinsac2012frequency,chernyakova2014fourier,david2015time}. Despite the promising results, the application of CS in US imaging still remains challenging. Firstly, we may remark that RIP cannot strictly hold in practical situations, mainly because of the lack of incoherence between the practical measurement and sparsity basis or because of the low level of sparsity of US images. As a consequence, the images reconstructed through CS are usually less good compared to standard acquisitions, especially when the compressive ratio (CS ratio) is low. In this paper, the CS ratio refers to the ratio between the number of linear measurements and the number of samples in the image to reconstruct. Secondly, the resolution of the reconstructed images is at most equivalent to those acquired using standard schemes. Nonetheless, it is well known that the spatial resolution, the signal-to-noise ratio and the contrast of standard US images are affected by the limited bandwidth of the imaging transducer, the physical phenomena related to US wave propagation such as diffraction and the imaging system. 

In order to overcome these issues, we have recently proposed a compressive deconvolution (CD) method aiming to reconstruct enhanced RF images from compressed linear measurements \cite{ChenCDTmi2015}. The main idea behind CD is to combine CS and deconvolution reconstructions into an unique framework leading to the following linear model:

\begin{equation}
\bsy = \Phi H\bsx + \bsn
\label{CsDec}
\end{equation}

where $\bsy\in \mathbb{R}^{M}$ contains $M$ linear measurements obtained by projecting one RF image $H\bsx\in \mathbb{R}^{N}$ onto the CS acquisition matrix $\Phi\in \mathbb{R}^{M\times N}$, with $M<<N$. $H \in \mathbb{R}^{N\times N}$ is a block circulant with circulant block (BCCB) matrix modelling the 2D convolution between the 2D PSF of the US system and the tissue reflectivity function (TRF) $\bsx\in \mathbb{R}^{N}$. In other words, the multiplication of the TRF by $H$ models the US RF image degradation mentioned above. Finally, $\bsn\in \mathbb{R}^{M}$ stands for a zero-mean additive white Gaussian noise. We emphasize that all the images in \eqref{CsDec} are expressed in the standard lexicographical order.

We should note that similar models have been recently proposed for general image processing purpose \cite{ma2009deblurring, xiao2011compounded, zhao2010compressed, amizic2013compressive,spinoulas2012simultaneous} including a theoretical derivation of RIP for random mask imaging \cite{7286775}. Nevertheless, in contrast to the solutions provided by these existing works, we showed in \cite{ChenCDTmi2015} that inverting \eqref{CsDec} by minimizing the following unconstrained objective function is well suitable for US imaging:

\begin{equation}
\hsx = \underset{x}{argmin}\parallel\Psi^{-1}H\bsx\parallel_1 + \alpha\parallel \bsx\parallel_p^p + \frac{1}{2\mu}\parallel\bsy-\Phi H\bsx\parallel_2^2
\label{Optmization}
\end{equation}

This objective function is composed by three terms: i) the $l_1$-norm term that aims at imposing the sparsity of the RF data $H\boldsymbol{x}$ in a transformed domain $\Psi$, ii) the $l_p$-norm ($1\leq p\leq 2$) regularizing the TRF $\boldsymbol{x}$ based on generalized Gaussian distribution (GGD) statistical assumption of US images ($p$ is related to the shape parameter of the GGD), see e.g. \cite{alessandrini2011restoration, zhao2014restoration,7174535}, iii) the data fidelity term. In order to solve the optimization problem in \eqref{Optmization}, the solution proposed in \cite{ChenCDTmi2015}  was based on the Alternative Direction Method of Multipliers (ADMM) \cite{boyd2011distributed}. 

In this paper, we further improve the US compressive deconvolution scheme in \cite{ChenCDTmi2015} by proposing a new reconstruction algorithm based on the Simultaneous Direction Method of Multipliers (SDMM) \cite{Setzer2010193}. Results on simulated and experimental images show improved convergence properties obtained with the proposed optimization routine, resulting into better reconstruction results and lower computational times compared to our previous work.  Moreover, we extend the CD approach to non-orthogonal measurement matrices, thus covering a more general compressed acquisition model.

This paper is organized as follows. We first recall the general framework of SDMM in section \ref{secGeneral}. The proposed SDMM-based optimization scheme able to solve \eqref{Optmization} is detailed in section \ref{secProposed}. In \ref{secResults}, simulated and experimental results are provided to show the effectiveness of the proposed method and its efficiency in recovering the TRF from compressed US data. The conclusions are drawn in Section \ref{secConcolusion}.

\section{General framework of Simultaneous Direction Method of Multipliers}
\label{secGeneral}
The algorithm of Simultaneous Direction Method of Multipliers (SDMM) e.g, \cite{Setzer2010193}, generalizes the alternating split Bregman method (ASB) \cite{goldstein2009split} to a sum of more than two functions. The ASB was initially proposed to solve optimization problems that can be expressed in the following form:

\begin{equation}
\underset{u\in\mathbb{R}^s,v\in\mathbb{R}^t}{argmin}\quad f(u)+g(v)\quad s.t.\quad v = Cu
\label{SplitBreg}
\end{equation}

where $C\in\mathbb{R}^{t\times s}$ is a given matrix, $f:\mathbb{R}^s\to\bar{\mathbb{R}}$ and $g:\mathbb{R}^t\to\bar{\mathbb{R}}$ are convex functions. $\bar{\mathbb{R}}$ is designated for extended real numbers, \textit{i.e.} $\mathbb{R}\bigcup\{+\infty\}$.

The iterative ASB method declines as follows:

\begin{equation}
u^{k+1}=\underset{u\in\mathbb{R}^s}{argmin}\quad f(u)+\frac{1}{2\beta}\parallel b^k+Cu-v^k\parallel_2^2
\label{SplitBreg1}
\end{equation}

\begin{equation}
v^{k+1}=\underset{v\in\mathbb{R}^t}{argmin}\quad g(v)+\frac{1}{2\beta}\parallel b^k+Cu^{k+1}-v\parallel_2^2
\label{SplitBreg2}
\end{equation}

\begin{equation}
b^{k+1} = b^k+Cu^{k+1}-v^{k+1}
\label{SplitBreg3}
\end{equation}

where $b\in\mathbb{R}^t$ is the Lagrangian parameter. It has been proven that the alternating split Bregman method is equivalent to Alternating Direction Method of Multipliers (ADMM) when the constraints are linear \cite{esser2009applications}.

Inspired from ASB, the general optimization problem considered in the framework of SDMM is:

\begin{equation}
\underset{u\in\mathbb{R}^s}{argmin}\sum_{i=1}^{m}f_i(C_iu)  
\label{SDMMGeneral}
\end{equation}

where $C_i\in\mathbb{R}^{t_i,s}$ and $f_i:\mathbb{R}^{t_i}\rightarrow\bar{\mathbb{R}}$ are convex functions. Considering $v_i\in\mathbb{R}^{t_i}$, $v_i=C_iu$, $f(u)=\left \langle 0,u \right \rangle$ and $g(v)=\sum_{i=1}^{m}f_i(v_i)$, \eqref{SDMMGeneral} can be reformulated as 

\begin{equation}
\underset{u\in\mathbb{R}^s,v_i\in\mathbb{R}^t_i}{argmin}\quad f(u)+\sum_{i=1}^{m}f_i(v_i)  
\label{SDMMGeneral2}
\end{equation}

Similarly to ASB method, SDMM iteratively solves the optimization problem above as follows:

\begin{equation}
u^{k+1} = \underset{u\in\mathbb{R}^s}{argmin}{\frac{1}{2\beta}\parallel\begin{pmatrix}b_1^k\\ \vdots \\ b_m^k\end{pmatrix}
+
\begin{pmatrix}C_1\\ \vdots \\ C_m\end{pmatrix}u
-
\begin{pmatrix}v_1^k\\ \vdots \\ v_m^k\end{pmatrix}
\parallel^2}
\label{SDMMGeneralStep1}
\end{equation}

\begin{equation}
\begin{pmatrix}v_1^{k+1}\\ \vdots \\ v_m^{k+1}\end{pmatrix}
= \underset{v_i\in\mathbb{R}^t_i}{argmin}
\{
\frac{1}{2\beta}
\parallel
\begin{pmatrix}b_1^k\\ \vdots \\ b_m^k\end{pmatrix}
+
\begin{pmatrix}C_1\\ \vdots \\ C_m\end{pmatrix}u^{k+1}
-
\begin{pmatrix}v_1\\ \vdots \\ v_m\end{pmatrix}\parallel^2
+\sum_{i=1}^mf_i(v_i)
\}
\label{SDMMGeneralStep2}
\end{equation}

\begin{equation}
\begin{pmatrix}b_1^{k+1}\\ \vdots \\ b_m^{k+1}\end{pmatrix} = \begin{pmatrix}b_1^k\\ \vdots \\ b_m^k\end{pmatrix}+\begin{pmatrix}C_1\\ \vdots \\ C_m\end{pmatrix}u^{k+1}-\begin{pmatrix}v_1^{k+1}\\ \vdots \\ v_m^{k+1}\end{pmatrix}
\label{SDMMGeneralStep3}
\end{equation}

\section{Proposed compressive deconvolution method}
\label{secProposed}

In this paper we propose an SDMM-based optimization scheme adapted to solve the problem in \eqref{Optmization}. First, we remark that \eqref{Optmization} can be reformulated as

\begin{equation}
\underset{\bsx}{argmin}\quad f_1(\bsv_1) + f_2(\bsv_2) + f_3(\bsv_3)
\label{CsDecSDMM}
\end{equation}

with

\[
\left\{\begin{matrix}
f_1(\bsv_1)=\alpha\parallel\bsv_1\parallel_p^p\\
f_2(\bsv_2)=\parallel\bsv_2\parallel_1 \\
 f_3(\bsv_3)=\frac{1}{2\mu}\parallel\bsy-\Phi \bsv_3\parallel_2^2
\\ 
\bsv_1 = C_1\bsx, \bsv_2 = C_2\bsx, \bsv_3 = C_3\bsx
\\ 
C_1 = I_N, C_2 = \Psi^{-1} H, C_3 = H
\end{matrix}\right.
\]

Using the parametrization above, the SDMM steps given in eq. \eqref{SDMMGeneralStep1}-\eqref{SDMMGeneralStep3} write for our compressive deconvolution problem as follows: 

\begin{equation}
\bsx^{k+1} = \underset{\bsx\in\mathbb{R}^N}{argmin}\quad\frac{1}{2\beta}\parallel\begin{pmatrix}\bsb_1^k\\ \bsb_2^k \\ \bsb_3^k\end{pmatrix}
+
\begin{pmatrix}I_N\\ \Psi^{-1}H \\ H\end{pmatrix}\bsx
-
\begin{pmatrix}\bsv_1^k\\ \bsv_2^k \\ \bsv_3^k\end{pmatrix}
\parallel^2
\label{SDMMstep1}
\end{equation}

\begin{equation}
\begin{pmatrix}\bsv_1^{k+1}\\ \bsv_2^{k+1} \\ \bsv_3^{k+1}\end{pmatrix}
= \underset{\bsv_1,\bsv_2,\bsv_3}{argmin}
\{
\frac{1}{2\beta}
\parallel
\begin{pmatrix}\bsb_1^k\\ \bsb_2^k \\ \bsb_3^k\end{pmatrix}
+
\begin{pmatrix} I_N \\\Psi^{-1}H\\ H\end{pmatrix}\bsx^{k+1}
-
\begin{pmatrix}\bsv_1\\ \bsv_2 \\ \bsv_3\end{pmatrix}\parallel^2
+\sum_{i=1}^3f_i(\bsv_i)
\}
\label{SDMMStep2}
\end{equation}

\begin{equation}
\begin{pmatrix}\bsb_1^{k+1}\\ \bsb_2^{k+1} \\ \bsb_3^{k+1}\end{pmatrix} = \begin{pmatrix}\bsb_1^k\\ \bsb_2^k \\ \bsb_3^k\end{pmatrix}+\begin{pmatrix}I_N \\ \Psi^{-1} H\\H\end{pmatrix}\bsx^{k+1}-\begin{pmatrix}\bsv_1^{k+1}\\ \bsv_2^{k+1} \\ \bsv_3^{k+1}\end{pmatrix}
\label{SDMMStep3}
\end{equation}

In the following, we give the details of solving each of the steps above. Firstly, we remark that Eq. \eqref{SDMMstep1} is a classical $l_2$-norm minimization problem that can be efficiently solved in the Fourier domain \cite{ng2010solving}.

Eq. \eqref{SDMMStep2} consists in solving three subproblems, corresponding to the update of $\bsv_1$, $\bsv_2$ and respectively $\bsv_3$. The $\bsv_1$-subproblem can be solved as follows:

\begin{equation}
\begin{split}
\bsv_1^{k+1} = &\underset{\bsv_1}{argmin}\quad\alpha\parallel\bsv_1\parallel_p^p+\frac{1}{2\beta}\parallel\bsb_1^k+\bsx^{k+1}-\bsv_1\parallel_2^2\\
= &prox_{\alpha\beta\parallel\cdot\parallel_p^p}(\bsb_1^k+\bsx^{k+1})
\end{split}
\label{SDMMstep2.1}
\end{equation}

where $prox$ represents the proximal operator \cite{pesquet2012parallel, pustelnik2011parallel, pustelnik2012relaxing}. The proximal operator of $\parallel \bsx\parallel_p^p$ has been given explicitly in the literature \cite{combettes2011proximal} and used in \cite{Chen2015.12}. More details about the proximal operator can be found in Appendix A.

The $\bsv_2$-subproblem can also be solved using the proximal operator associated to the $\ell_1$-norm that corresponds to the soft thresholding operator \cite{ng2010solving} (see Appendix A):

\begin{equation}
\begin{split}
\bsv_2^{k+1} = &\underset{\bsv_2}{argmin}\quad\parallel\bsv_2\parallel_1+\frac{1}{2\beta}\parallel\bsb_2^k+\Psi^{-1} H\bsx^{k+1}-\bsv_2\parallel_2^2\\
= &prox_{\beta\parallel\cdot\parallel_1}(\bsb_2^k+\Psi^{-1} H\bsx^{k+1})
\end{split}
\label{SDMMstep2.2}
\end{equation}

Finally, the $\bsv_3$-subproblem can be solved as follows:

\begin{equation}
\begin{split}
\bsv_3^{k+1} & = \underset{\bsv_3}{argmin}\frac{1}{2\mu}\parallel y-\Phi\bsv_3\parallel_2^2+\frac{1}{2\beta}\parallel\bsb_3^k+H\bsx^{k+1}-\bsv_3\parallel_2^2\\
&\Leftrightarrow [\beta\Phi^t\Phi+\mu]\bsv_3^{k+1} = \beta\Phi^t\bsy+\mu\bsb_3^k+\mu H\bsx^{k+1}
\end{split}
\label{SDMMstep2.3}
\end{equation}

For orthogonal sampling matrices $\Phi$, the Sherman-Morrison-Woodbury inversion matrix lemma \cite{deng2013group} allows us to efficiently find the solution of the $\bsv_3$-subproblem above \cite{Chen2015.12}. In order to make our compressive deconvolution method more general and therefore relevant to various compressive acquisition schemes in US imaging, we give in Appendix B a solution of \eqref{SDMMstep2.3} for non orthogonal $\Phi$ matrices. 

Algorithm 1 summarizes the SDMM-based numerical scheme proposed for solving \ref{Optmization}.

\alglanguage{pseudocode}
\begin{algorithm}
\caption{Compressive deconvolution SDMM-based algorithm.}
\label{Algorithm:SDMM}
\begin{algorithmic}[1] 
	\Require $\alpha$, $\mu$, $\beta$, $\bsv_i^0$, $\bsb_i^0$, $i=1,2,3$ 
 	\While{not converged}
 	\State $\bsx^{k+1} \gets \bsv_i^k, \bsb_i^k$ \Comment{update $\bsx^{k+1}$ using \eqref{SDMMstep1}}
 	\State $\bsv_1^{k+1} \gets \bsb_1^k, \bsx^{k+1}$\Comment{update $\bsv_1^{k+1}$ using \eqref{SDMMstep2.1}}
 	\State $\bsv_2^{k+1} \gets \bsb_2^k, \bsx^{k+1}$\Comment{update $\bsv_2^{k+1}$ using \eqref{SDMMstep2.2}}
 	\State $\bsv_3^{k+1} \gets \bsb_3^k, \bsx^{k+1}$\Comment{update $\bsv_3^{k+1}$ using \eqref{SDMMstep2.3}}
 	\State $\bsb_i^{k+1} \gets \bsv_i^{k+1}, \bsx^{k+1}$\Comment{update $\bsb_i^{k+1}$ using \eqref{SDMMStep3}}
    \EndWhile  
    \Ensure $\bsx$ 
\end{algorithmic}
\end{algorithm}

We emphasize that compared to the ADMM-based scheme that we have recently proposed to solve \eqref{Optmization} \cite{ChenCDTmi2015}, the method resumed in Algo. \ref{Algorithm:SDMM} requires one less hyperparameter. Moreover, with the proposed optimization scheme all the subproblems are solved exactly, while in \cite{ChenCDTmi2015} we have only obtained an approximation for the $\bsv_1$-subproblem in eq. \eqref{SDMMstep2.1}.  This improvement allows the SDMM-based iterative scheme to converge faster than the ADMM-based algorithm proposed in \cite{ChenCDTmi2015}. Since this $\bsv_1$-subproblem is critical for the deconvolution process, one may also expect more accurate compressive deconvolution results with SDMM than with ADMM.

\section{Simulation Results}
\label{secResults}
In this section, we provide numerical experiments to evaluate the effectiveness of the proposed compressive deconvolution optimization framework, denoted by SDMM hereafter. Since we have recently shown in \cite{ChenCDTmi2015} the superiority of the ADMM-based method (denoted by ADMM in this section) compared to other compressive deconvolution methods, the technique in \cite{ChenCDTmi2015} is used herein for comparison purpose \footnote{The code corresponding to the ADMM-based method is available \url{http://www.irit.fr/~Adrian.Basarab/codes.html}}. 

\subsection{Results on simulated data}
\label{simuresults}
\begin{figure*}[!t]
\centering
\includegraphics[scale=0.26]{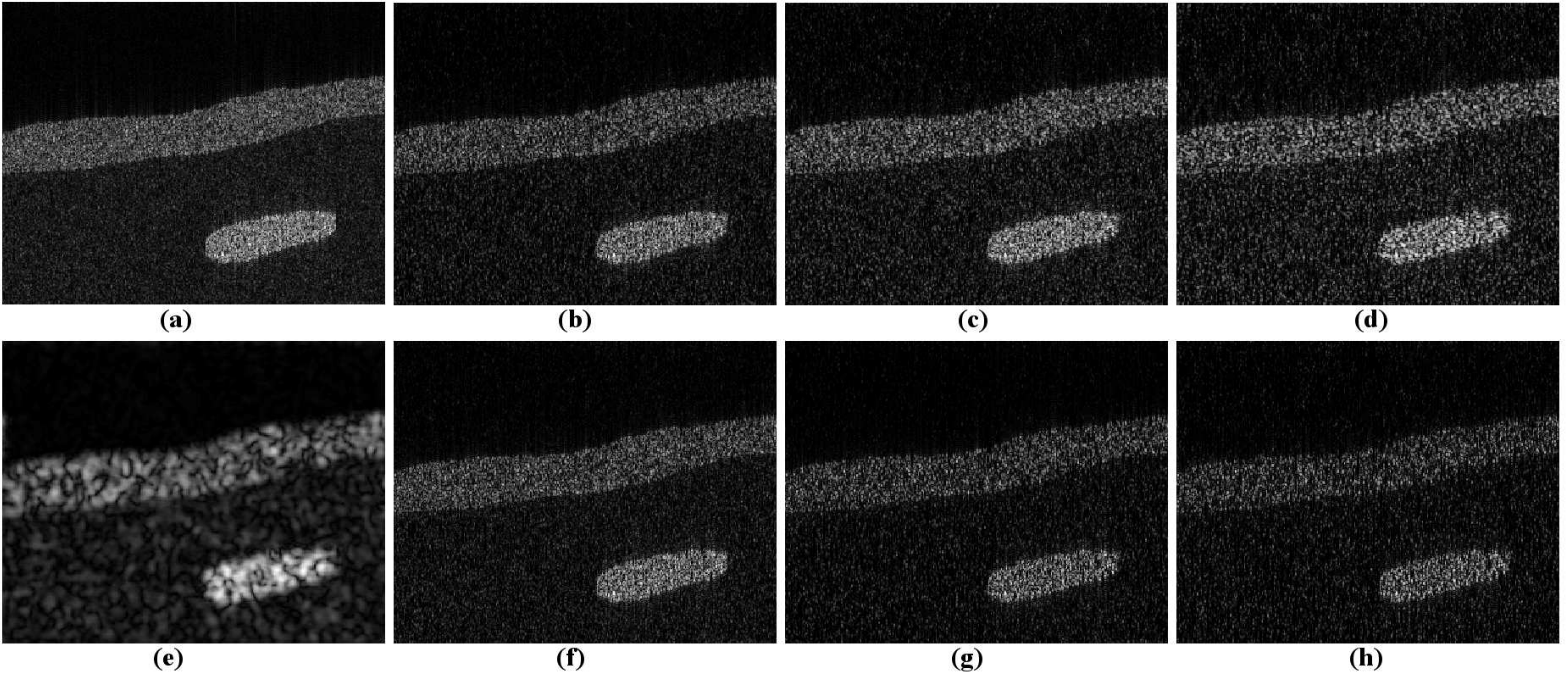}
\caption{Results on simulated data (Group 1). (a) TRF, (b-d) Reconstruction results using ADMM for CS ratios of 0.6, 0.4 and 0.2, (e) Simulated US image, (f-h) Reconstruction results using SDMM for CS ratios of 0.6, 0.4 and 0.2.}
\label{fig_cartoon}
\end{figure*}

Two groups of simulation experiments (named Group 1 and 2) have been conducted to evaluate the performance of the proposed scheme. The RF images have been generated following the procedure in \cite{ng2007wavelet} using a 2D convolution between a US PSF and a map of scatterers, i.e, tissue reflectivity function (TRF). In both cases, the PSF was generated using a Field II \cite{jensen1991model} simulation corresponding to a 128-element linear probe operating at 3.5 MHz and an axial sampling frequency of 20 MHz.
The TRFs were generated using an well-established procedure in US imaging, by assigning the scatterers random amplitudes following a given distribution, weighted by a cartoon image denoted by mask hereafter. For the first group, a Laplacian distribution has been employed and the mask has been hand drawn to simulate four different regions with different echogenicity. The resulting TRF and US image (plotted in B-mode) are shown in     
Fig. \ref{fig_cartoon} (a) and (e).

\begin{figure*}[!t]
\centering
\includegraphics[scale=0.26]{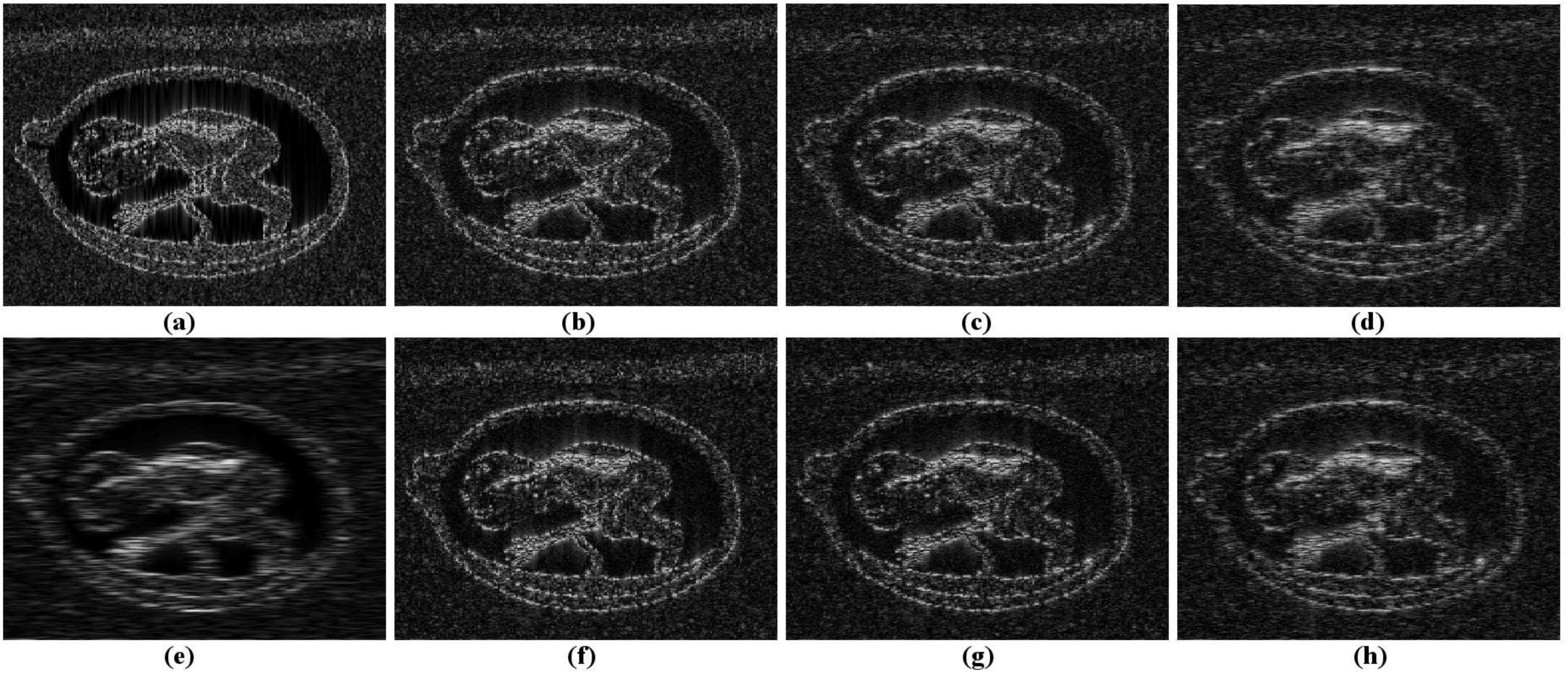}
\caption{Results on simulated data (Group 2). (a) TRF, (b-d) Reconstruction results using ADMM for CS ratios of 0.6, 0.4 and 0.2, (e) Simulated US image, (f-h) Reconstruction results using SDMM for CS ratios of 0.6, 0.4 and 0.2.}
\label{fig_fetus}
\end{figure*}

The TRF of Group 2 follows one of the examples proposed by the Field II simulator \cite{jensen1991model}, mimicking a fetus. In this case, the scatterer amplitudes were generalized Gaussian distributed, with the shape parameter of the GGD equal to $1.5$. The TRF and the simulated US image are displayed in Fig. \ref{fig_fetus} (a) and (e). For both simulations, the number of scatterers was considered sufficiently large ($10^5$) to ensure fully developed speckle.

The compressed measurements were obtained in the same manner for both groups, by projecting the RF images onto orthogonal Structurally Random Matrix (SRM) \cite{do2012fast} and were degraded by an additive Gaussian noise corresponding to a SNR of 40 dB. We emphasize that results corresponding to a non-orthogonal measurement matrix are also provided in the Appendix B.

Fig. \ref{fig_cartoon} and Fig. \ref{fig_fetus} display the compressive deconvolution reconstruction results obtained with ADMM \cite{ChenCDTmi2015} and with the proposed SDMM scheme for CS ratios of 0.6, 0.4 and 0.2. The value of $p$ used to regularize the TRF estimations was set to 1 for Group 1 and 1.5 for Group 2. All the other hyperparameters were set to their best possible values by cross-validation. We should note that since both ADMM and SDMM methods aim at solving the same objective function in \eqref{Optmization}, the hyperparameters $\alpha$ and $\mu$ have been assigned the same values in order to ensure a fair comparison. For the same reason, both algorithms were assigned the same convegence criterion, \textit{i.e.} $\parallel \boldsymbol{x}^k - \boldsymbol{x}^{k-1} \parallel/\parallel \boldsymbol{x}^{k-1}\parallel<5e^{-4} $, with $k$ the iteration number and $\bsx_k$ the estimated image at iteration $k$. 

Taking benefit from the fact that the TRF ground truth is available in simulation experiments, the peak signal-to-noise ratio (PSNR) and the structural similarity (SSIM) are used in this paper to assess the quality of the reconstruction results. Higher PSNR or SSIM indicates that the reconstruction is of higher quality. PSNR is usually expressed in terms of the logarithmic decibel scale and defined as

\begin{equation}
PSNR = 10log_{10}\frac{NL^2}{\bsx-\hat{\bsx}}
\end{equation}

where $\bsx$ and $\hat{\bsx}$ are the original and reconstructed images, and the constant $L$ represents the maximum intensity value in $\bsx$. SSIM is usually measured in percentage and defined as

\begin{equation}
SSIM = \frac{(2\mu_\bsx\mu_{\hat{\bsx}}+c_1)(2\sigma_{\bsx\hat{\bsx}}+c_2)}{(\mu_\bsx^2+\mu_{\hat{\bsx}}^2+c_1)(\sigma_\bsx^2+\sigma_{\hat{\bsx}}^2+c_2)}
\label{eq28}
\end{equation}

where $\bsx$ and $\hat{\bsx}$ are the original and reconstructed images, $\mu_\bsx$, $\mu_{\hat{\bsx}}$, $\sigma_\bsx$ and $\sigma_{\hat{\bsx}}$ are the mean and variance values of $\bsx$ and $\hat{\bsx}$, $\sigma_{\bsx\hat{\bsx}}$  is the covariance between $\bsx$ and $\hat{\bsx}$; $c_1=(k_1C)^2$ and $c_2=(k_2C)^2$ are two variables aiming at stabilizing the division with weak denominator, $C$ is the dynamic range of the pixel-values and $k_1$, $k_2$ are constants. Herein, $C=1$, $k_1=0.01$ and $k_2=0.03$.

These quantitative results are regrouped in Table.\ref{tab:simu}, where bold values stand for the best result obtained for each experiment.

\begin{table}[htbp]
  \centering
  \caption{Quantitative results for compressive deconvolution reconstruction of simulated US images}
\begin{tabular}{c|c|cccc}
    \hline
    & CS ratios &0.8 & 0.6 & 0.4 & 0.2\\
    \hline
    \multicolumn{6}{c}{Group 1}\\
    \hline
    \multirow{2}[0]{*}{ADMM} &  PSNR(dB) & 29.14 & 28.34 & 27.01 & 24.60\\
    &  SSIM(\%) & 81.58 & 77.44 & 69.07 & 51.65\\
    \hline
	\multirow{2}[0]{*}{SDMM} &  PSNR(dB) & \textbf{30.67} & \textbf{29.55} & \textbf{27.94} & \textbf{26.18}\\
    &  SSIM(\%) & \textbf{85.77} & \textbf{81.66} & \textbf{74.37} & \textbf{63.15}\\
    \hline
    \multicolumn{6}{c}{Group 2}\\
    \hline
    \multirow{2}[0]{*}{ADMM} &  PSNR(dB) & 24.38 & 22.67 & 20.70 & 18.54\\
    &  SSIM(\%) & 63.31 & 53.63 & 39.04 & 19.96\\
    \hline
	\multirow{2}[0]{*}{SDMM} &  PSNR(dB) & \textbf{25.07} & \textbf{22.97} & \textbf{20.77} & \textbf{18.57}\\
    &  SSIM(\%) & \textbf{64.48} & \textbf{54.53} & \textbf{39.39} & \textbf{20.05}\\
    \hline
        \end{tabular}%
  \label{tab:simu}%
\end{table}%

\begin{figure*}[!t]
\centering
\includegraphics[scale=0.3]{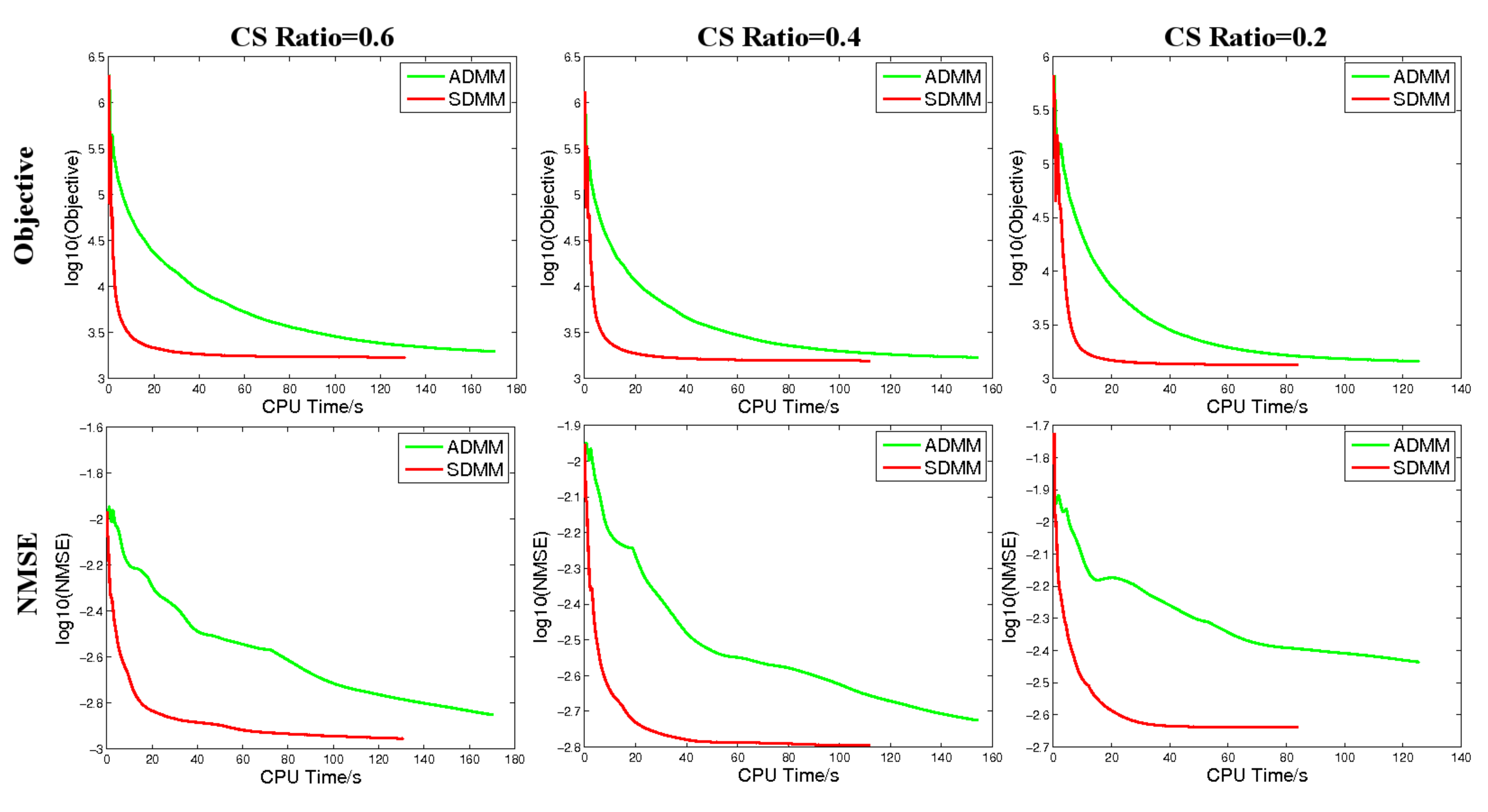}
\caption{Convergence performance on simulated data (Group1).}
\label{fig_cartoon_converge}
\end{figure*}

\begin{figure*}[!t]
\centering
\includegraphics[scale=0.4]{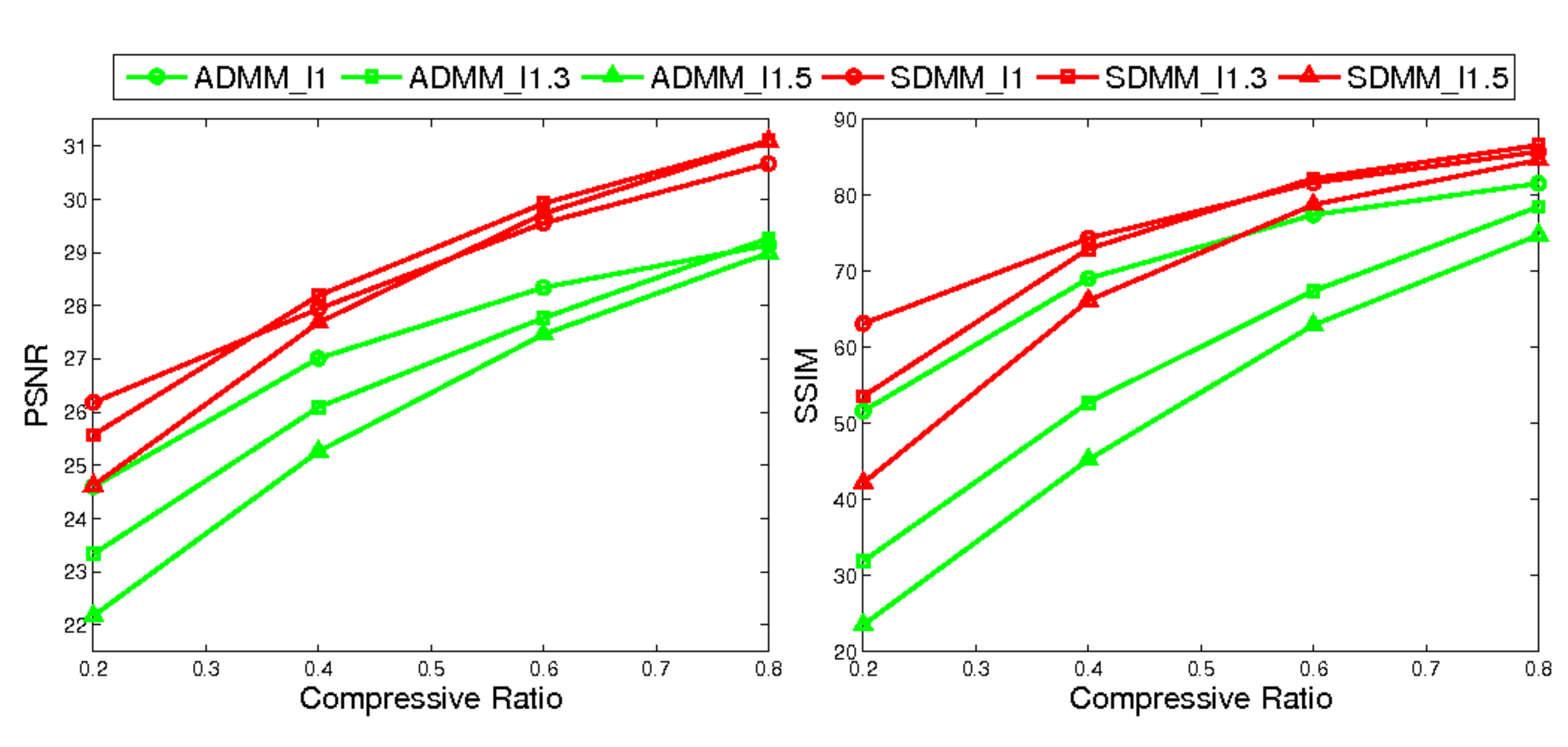}
\caption{Results of all the methods with different $p$ on simulated data (Group1).}
\label{fig_lpResults}
\end{figure*}

Both the visual inspection of images in figures \ref{fig_cartoon} and \ref{fig_fetus} and the quantitative results in Table.\ref{tab:simu} show that the proposed SDMM-based method outperforms the ADMM algorithm for the two simulated images and for all the CS ratios. In addition to the reconstruction quality gain, the proposed method also offers better convergence properties compared to ADMM. This convergence improvement is clearly highlighted by the plots in Fig. \ref{fig_cartoon_converge}. We may thus remark that for all the CS ratios, the convergence curves, both in terms of objective function (as eq. \eqref{Optmization}) and of Normalized Mean Square Error (NMSE) defined in eq.\eqref{NMSEdef}, decreases much faster with SDMM than with ADMM. The computations were performed using a computer with Intel Xeon CPU E5620 @2.40GHz, 4.00G RAM. Depending on the stopping criterion, the convergence rate of SDMM for Group 1 is at least twice faster than the one of ADMM. We emphasize that the same convergence properties have been obtained for Group 2.

\begin{equation}
NMSE = \frac{1}{N}\parallel\bsx-\hat{\bsx}\parallel^2_2
\label{NMSEdef}
\end{equation}

where $\bsx$ and $\hat{\bsx}$ are the normalized original and reconstructed TRF images and N represents the number of pixels in the image.

As explained previously, the value of the regularization parameter $p$ has been manually tuned in the two simulated experiments. However, one may notice the importance of this parameter on the reconstruction results, as it directly affects the regularization of the TRF \cite{7174535}. In order to show its influence on the results, we regroup in Fig. \ref{fig_lpResults} the PSNR and SSIM results for both SDMM and ADMM methods for three values of $p$, versus the CS ratio. In addition to the superiority of SDMM compared to ADMM, one may remark that the choice of $p$ is more important for low CS ratios. This observation can be explained by the further importance of the regularization when only a small amount of data is available.

\subsection{Results on \textit{in vivo} data}
\label{invivoresults}

\begin{figure*}[!t]
\centering
\includegraphics[scale=0.26]{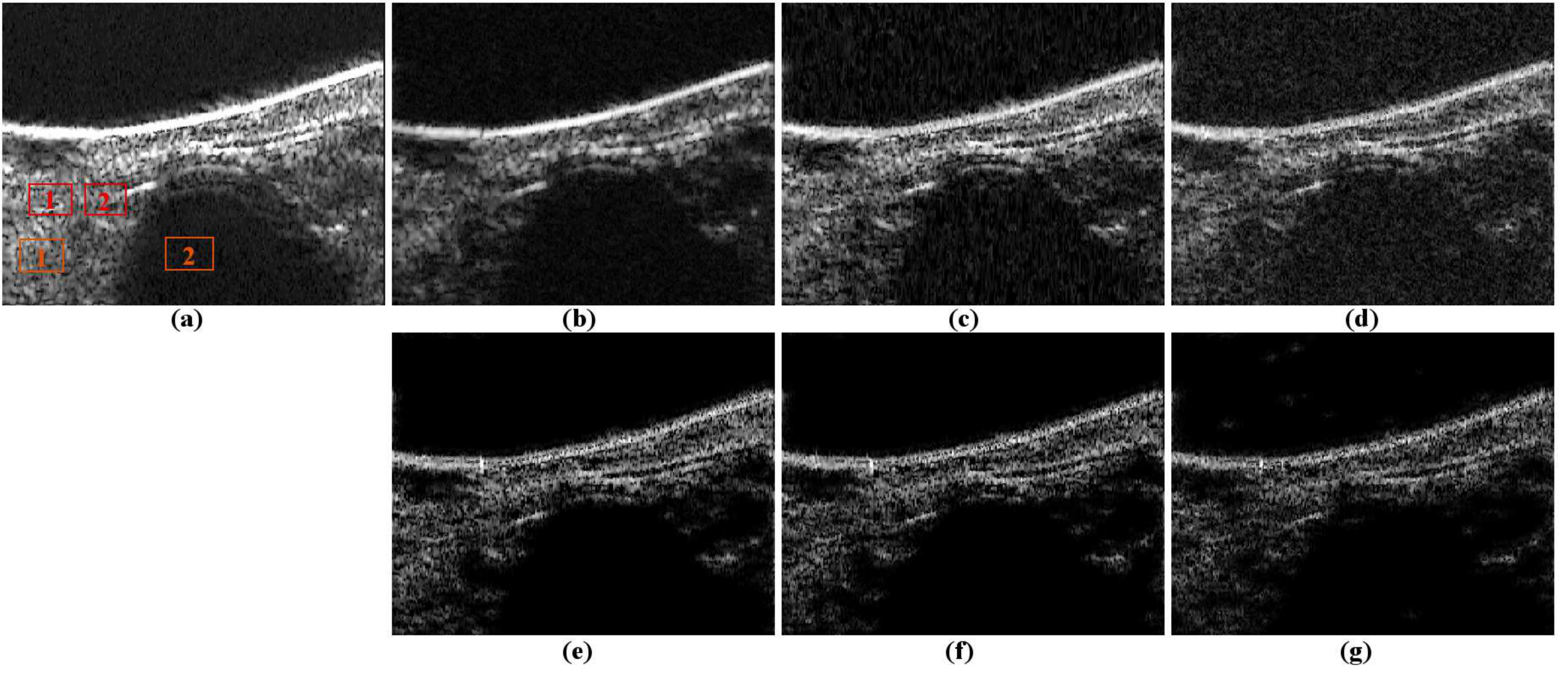}
\caption{Results on \textit{in vivo} data (Group 3). (a) Original US image, (b-d) Reconstruction results using ADMM for CS ratios of 0.8, 0.6 and 0.4, obtained for $p=1$, (e-g) Reconstruction results using SDMM for CS ratios of 0.8, 0.6 and 0.4, obtained for $p=1$.}
\label{fig_MouseBladder}
\end{figure*}

\begin{figure*}[!t]
\centering
\includegraphics[scale=0.26]{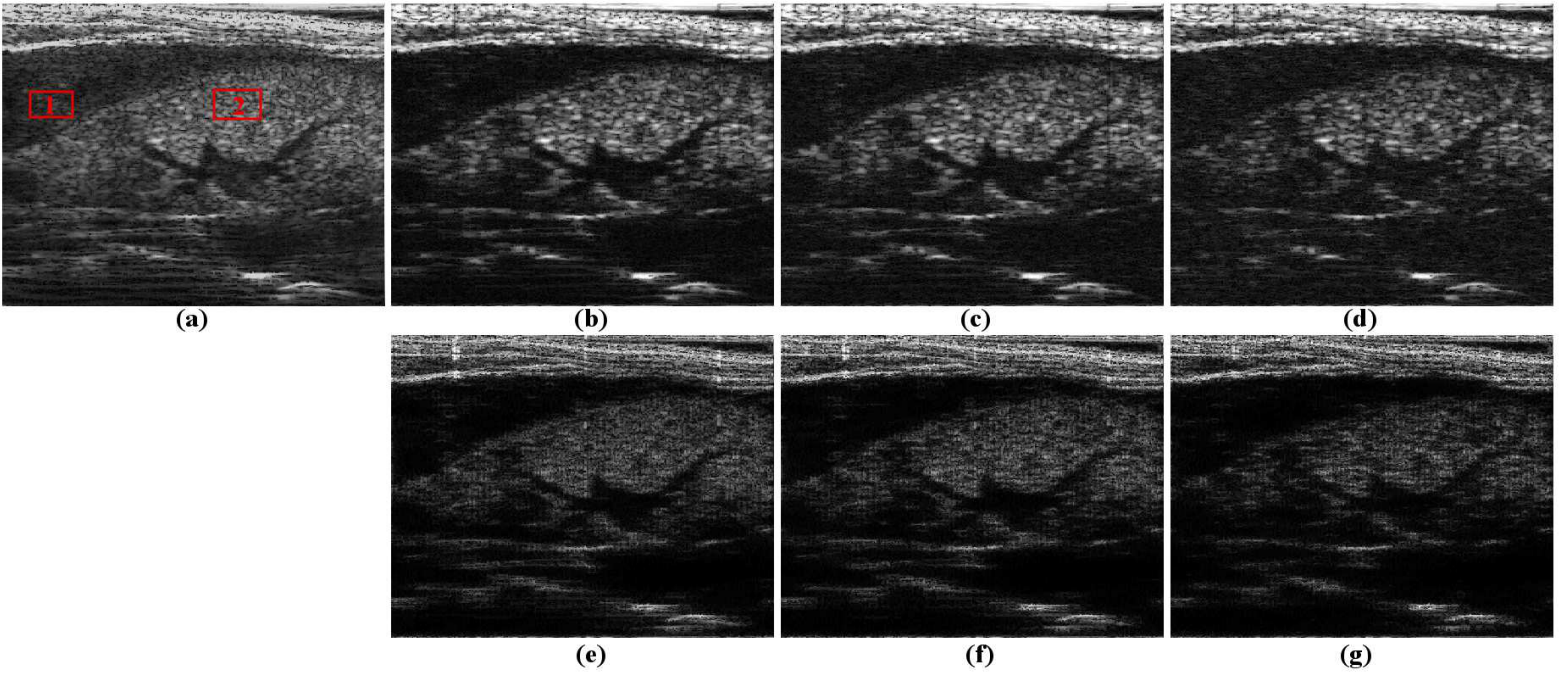}
\caption{Results on \textit{in vivo} data (Group 4). (a) Original US image, (b-d) Reconstruction results using ADMM for CS ratios of 0.8, 0.6 and 0.4, obtained for $p=1.5$, (e-g) Reconstruction results using SDMM for CS ratios of 0.8, 0.6 and 0.4, obtained for $p=1.5$.}
\label{fig_Mousekidney}
\end{figure*}

In this section, we evaluate the results of the proposed SDMM-based compressive deconvolution method on two \textit{in vivo} US images, denoted by Group 3 and Group 4. Group 3 corresponds to a mouse bladder shown in Fig. \ref{fig_MouseBladder} (a), while Group 4 represents a mouse kidney, see Fig. \ref{fig_Mousekidney} (a). Both images were acquired with a 20 MHz single-element US probe. Since the PSF is unknown in practical situations, it has been initially estimated from the data, as a pre-processing step, following the PSF estimation procedure presented in \cite{michailovich2005novel}. The compressive deconvolution results obtained with ADMM and SDMM are shown in figures \ref{fig_MouseBladder} (b-g) and \ref{fig_Mousekidney} (b-g) for CS ratios of 0.8, 0.6 and 0.4. Given the "sparse" appearance of the mouse bladder caused by the weak amount of scatterers in the liquid, the value of $p$ was set to 1 for Group 3 and to 1.5 for Group 4.

For the \textit{in vivo} data, the true TRFs are obviously not available, making thus impossible the computation of quantitative results such as the PSNR or the SSIM. As a consequence, the quality of the compressive deconvolution results is evaluated in this section according to the standard contrast-to-noise ratio (CNR). Moreover, CPU times for both ADMM and SDMM reconstructions are shown in Table \ref{tab:invivo}. The CNR values were computed for the regions highlighted by the red or orange rectangles in Figures \ref{fig_MouseBladder} and \ref{fig_Mousekidney}. For instance, two CNRs have been calculated for Group 3, between one region in the bladder cavity and respectively two regions extracted from the bladder wall. The results in Table \ref{tab:invivo} show equivalent results between ADMM and SDMM. Nevertheless, SDMM was roughly 2 to 6 times faster than ADMM, due to its better convergence properties discussed in the previous section.

The visual inspection of the results highlights better denoising achievements with SDMM compared to ADMM, as for example in weak scatterer regions such as the bladder cavity. We emphasize that the reconstructed TRF in Figures \ref{fig_MouseBladder} and \ref{fig_Mousekidney} are shown after envelope detection and log compression, in order to be comparable to the standard B-mode images. However, the deconvolution process results into TRFs that, contrary to RF images, are not longer modulated in the axial direction. Indeed, the carrier information is included in the PSF that is eliminated during the deconvolution process. For this reason, the standard procedure of envelope detection based on the amplitude of the complex analyitic signal, is not adapted to TRF. Instead, we have used an envelope estimator based on the detection and interpolation of local maximum, classically used in empirical mode decomposition techniques \cite{flandrin2004empirical}. 

\begin{table}[htbp]
  \centering
  \caption{Quantitative results for the \textit{in vivo} data}
\begin{tabular}{c|c|cccc}
    \hline
    & CS ratios & 1 & 0.8 & 0.6 & 0.4 \\
    \hline
    \multicolumn{6}{c}{Group 3}\\
    \hline
    \multirow{3}[0]{*}{ADMM} &  CNR1 & \textbf{1.65} & \textbf{1.63} & \textbf{1.57} & \textbf{1.40}\\
    & CNR2 & \textbf{2.51} & 2.00 & 1.52 & 1.10\\
    &  Time(s) & 76.40 & 77.36 & 100.88 & 112.96\\
    \hline
	\multirow{3}[0]{*}{SDMM} &  CNR1 & 1.40 & 1.39 & 1.38 & 1.23\\
	& CNR2 & 2.33 & \textbf{2.29} & \textbf{2.05} & \textbf{1.50}\\
    &  Time(s) & \textbf{16.40} & \textbf{18.71} & \textbf{24.76} & \textbf{41.31}\\
    \hline
    \multicolumn{6}{c}{Group 4}\\
    \hline
    \multirow{2}[0]{*}{ADMM} &  CNR & \textbf{1.98} & \textbf{1.90} & 1.77 & 1.40\\
    &  Time(s) & 629.57 & 561.20 & 484.93 & 343.09\\
    \hline
	\multirow{2}[0]{*}{SDMM} &  CNR & 1.90 & 1.87 & \textbf{1.84} & \textbf{1.58}\\
    &  Time(s) & \textbf{186.64} & \textbf{216.66} & \textbf{265.43} & \textbf{312.12}\\
    \hline
        \end{tabular}%
  \label{tab:invivo}%
\end{table}%

\section{Conclusion}
\label{secConcolusion}
Reconstructing enhanced US images from compressed measurements is a very recent paradigm that regroups compressive sampling and deconvolution into a sole framework. The main objective of this paper was to propose an SDMM-based algorithm dedicated to solve the compressive deconvolution problem in US imaging. Compared to an ADMM-based method that we have recently published in \cite{ChenCDTmi2015}, the proposed algorithm requires one less hyperparameter since one of the optimization subproblems can be solved without any approximation. Moreover, the proposed variable splitting scheme made possible by SDMM is shown to allow faster convergence compared to ADMM. Finally, an alternative to compressed measurements obtained with non-orthogonal matrices is provided, thus extending the practical interest of the compressive deconvolution approach. Our future work will include the consideration of blind deconvolution techniques able to jointly estimate the PSF and tissue reflectivity function, through statistical regularization techniques or parametric models. Moreover, an automatic choice of the optimal value of the regularization parameter $p$ would be of great interest in practice. This optimal choice may be considered through statistical assumptions on the US images, such as the heavy-tailed distributions discussed in \cite{7174535}. Finally, an interesting future researh track will be to evaluate the compressive deconvolution with specific compressed measurements, such as those obtained by Xampling \cite{chernyakova2014fourier} or with optimized sparse arrays \cite{DIAR-13a}.

\appendix[Proximal operator]
The proximal operator (or proximal mapping) of a function $f$, denoted by $prox_f$, is defined by 

\begin{equation}
prox_f(x) = \underset{u\in\mathbb{R}^N}{argmin}\quad f(u)+\frac{1}{2}\parallel u-x\parallel_2^2
\label{proximal}
\end{equation}

When $f(u)=K\mid u\mid^p$, \eqref{proximal}

\begin{equation}
prox_{K\mid\cdot\mid^p}(x) = \underset{u}{argmin}\quad K\mid u\mid^p+\frac{1}{2}\parallel u-x\parallel_2^2
\end{equation}

or

\begin{equation}
prox_{K\mid\cdot\mid^p}(x) = \underset{u}{argmin}\quad \mid u\mid^p+\frac{1}{2K}\parallel u-x\parallel_2^2
\end{equation}

The unique solution to the minimization problem above given by \cite{pustelnik2011parallel} is:

\begin{equation}
prox_{K\mid\cdot\mid^p}(x)=sign(x)q
\end{equation}

where $q\geq 0$ and 

\begin{equation}
q + pKq^{p-1}=\mid x\mid
\end{equation}

\begin{figure}[!ht]
\centering
\includegraphics[scale=0.5]{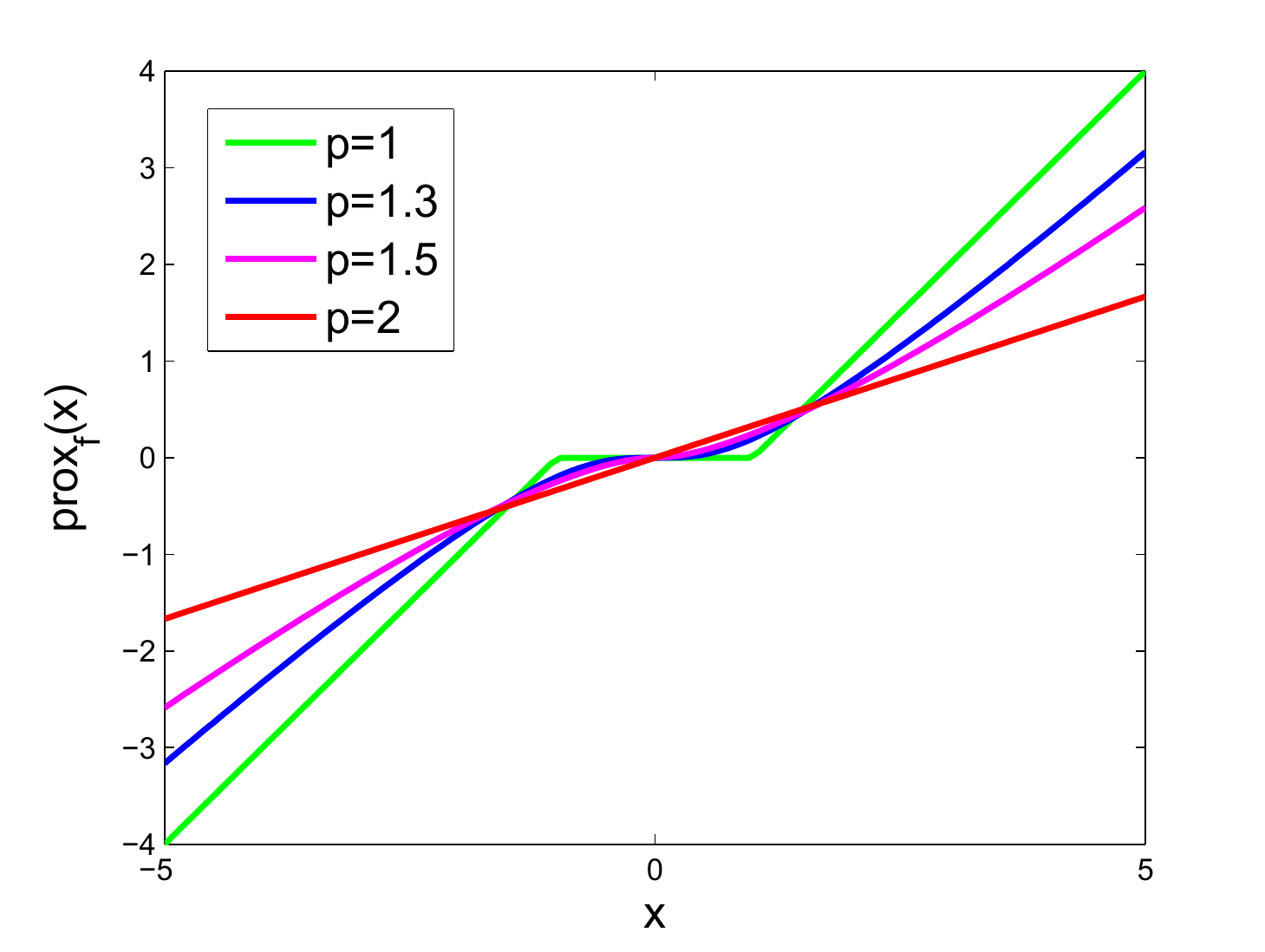}
\caption{Proximal operator of $\mid x\mid^p$ for different values of $p$}
\label{fig_PROXIMAL}
\end{figure}

For the case $p=1$, the proximal operator of $K\mid x\mid$ is the well known thresholding.
For the case $p\neq 1$, the numerical solution to the equation above, \textit{i.e.} the value of $q$, can be obtained using Newton's method. The resulting proximal operators for different values of $p$ are plotted in Fig.\ref{fig_PROXIMAL}.

\appendix[SDMM-based compressive deconvolution with non-orthogonal measurement matrices]

When the sampling matrix $\Phi$ is non-orthogonal, the solution of $\bsv_3$-subproblem in eq. \eqref{SDMMstep2.3} cannot be computed in practical situations because of the high-dimensional matrices. To overcome this issue, we propose to use Newton's method to approximate its solution.

Let us denote

\begin{equation}
h(\bsv_3)=[\beta\Phi^t\Phi+\mu]\bsv_3 -\beta\Phi^t\bsy+\mu\bsb_3^k+\mu H\bsx^{k+1}
\label{v3}
\end{equation}

At each iteration, we approximate $\bsv_3^{k+1}$ by

\begin{equation}
\bsv_3^{k+1} = \bsv_3^k - stp*h(\bsv_3^k)
\label{v3k}
\end{equation}

where $stp$ is defined as

\begin{equation}
stp = \frac{h(\bsv_3^k)^th(\bsv_3^k)}{\beta[\Phi h(\bsv_3^k)]^t[\Phi h(\bsv_3^k)]+\mu h(\bsv_3^k)^th(\bsv_3^k)}
\label{v3stp}
\end{equation}

To conclude, the solution to eq.\eqref{SDMMstep2.3} are shown in Algorithm \ref{Algorithm:appendix}.

\alglanguage{pseudocode}
\begin{algorithm}
\caption{Solution to eq.\eqref{SDMMstep2.3}}
\label{Algorithm:appendix}
\begin{algorithmic}[1] 
	\Require $\mu$, $\beta$, $\Phi$, $\bsy$, $H$, $\bsx^{k+1}$, $\bsv_3^k$, $\bsb_3^k$,  
 	\If{$\Phi$ is orthogonal}
 	\State{Update $\bsv_3^{k+1}$ by Sherman-Morrison-Woodbury inversion matrix lemma} 
 	\Else 
 	\State{Update $\bsv_3^{k+1}$ using eq.\eqref{v3k} }
 	\EndIf
\end{algorithmic}
\end{algorithm}

In order to evaluate the effectiveness of this method for non-orthogonal measurements matrices (the resulting SDMM scheme is denoted by nSDMM), we use hereafter the simulated image in Fig.\ref{fig_cartoon}. The compressive measurements are obtained by projecting the TRF onto a random Gaussian matrix. The reconstruction results with nSDMM are compared to those obtained in Section \ref{simuresults} with SDMM using an orthogonal measurement matrix. 

\begin{figure*}[!t]
\centering
\includegraphics[scale=0.26]{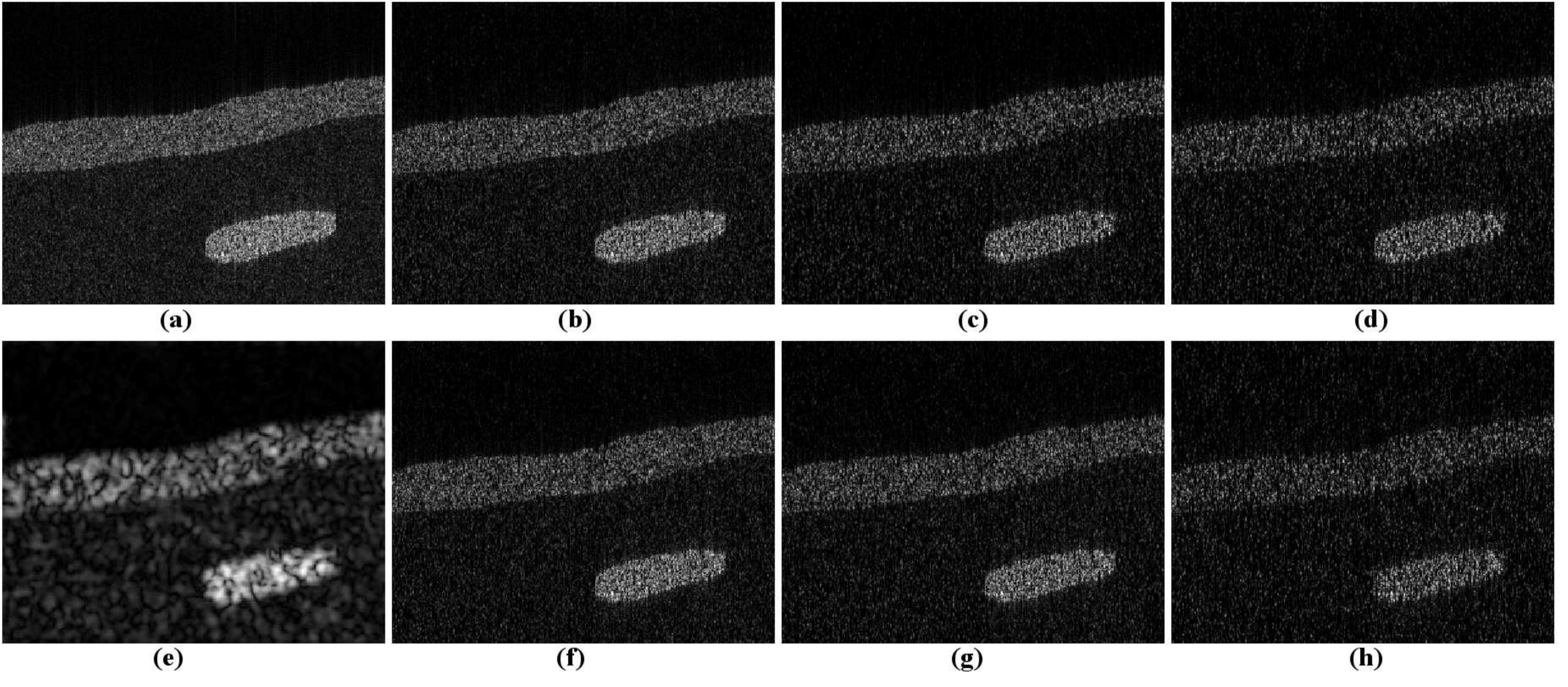}
\caption{Results on simulated data (Group 1). (a) TRF, (b-d) Reconstruction results using SDMM from orthogonal measurements for CS ratios of 0.6, 0.4 and 0.2, (e) Simulated US image, (f-h) Reconstruction results using nSDMM from non-orthogonal measurements for CS ratios of 0.6, 0.4 and 0.2.}
\label{fig_NSDMM}
\end{figure*}

\begin{table}[!ht]
  \centering
  \caption{Quantitative comparison between SDMM and nSDMM}
\begin{tabular}{c|c|cccc}
    \hline
    & CS ratios &0.8 & 0.6 & 0.4 & 0.2\\
    \hline
	\multirow{2}[0]{*}{SDMM} &  PSNR & \textbf{30.67} & \textbf{29.55} & 27.94 & \textbf{26.18}\\
    &  SSIM & \textbf{85.77} & \textbf{81.66} & \textbf{74.37} & \textbf{63.15}\\
    \hline
    \multirow{2}[0]{*}{nSDMM} &  PSNR & 30.33 & 29.36 & \textbf{28.02} & 26.00\\
    &  SSIM & 83.90 & 79.89 & 73.44 & 57.45\\
    \hline
        \end{tabular}%
  \label{tab:NSDMM}%
\end{table}%

Fig. \ref{fig_NSDMM} regroups the nSDMM and SDMM reconstructions for CS ratios of 0.6, 0.4 and 0.2, while the quantitative results are reported in Table. \ref{tab:NSDMM}. Despite a slight reconstruction degradation caused by the Newton approximation, one may remark that nSDMM is able to fairly reconstruct the TRF from non-orthogonal measurements.


\section*{Acknowledgment}

The authors are grateful to Prof. Jean-Christophe Pesquet at Universit\'{e} Paris-Est for his constructive suggestions.

\ifCLASSOPTIONcaptionsoff
  \newpage
\fi

\bibliographystyle{IEEEtran}
\bibliography{BibCsDecSDMM}

\end{document}